# Characterizing the Effect of Sentence Context on Word Meanings: Mapping Brain to Behavior

**Nora Aguirre-Celis (naguirre@cs.utexas.edu)**
ITESM, Monterrey, Mexico & The University of Texas at Austin
Department of Computer Science, 2317 Speedway
Austin, TX 78712 USA

**Risto Miikkulainen (risto@ cs.utexas.edu)**
The University of Texas at Austin
Department of Computer Science, 2317 Speedway
Austin, TX 78712 USA

## Abstract

Semantic feature models have become a popular tool for prediction and interpretation of fMRI data. In particular, prior work has shown that differences in the fMRI patterns in sentence reading can be explained by context-dependent changes in the semantic feature representations of the words. However, whether the subjects are aware of such changes and agree with them has been an open question. This paper aims to answer this question through a human-subject study. Subjects were asked to judge how the word change from their generic meaning when the words were used in specific sentences. The judgements were consistent with the model predictions well above chance. Thus, the results support the hypothesis that word meaning change systematically depending on sentence context.

## Introduction

Semantic feature theory suggests that a word meaning is instantiated by weighting its semantic attributes according to the context. (Barclay, et.al., 1974; Hampton, 1996; Kiefer & Pulvermüller 2012; Medin & Shoben, 1988; Mitchell & Lapata, 2010; Murphy, 1990; Wisniewsky, 1998). For example, when people think of the word *football*, they heavily weigh features like 'shape' and 'lower limbs' and features like 'smell' and 'size' lightly. In contrast, when they think of *forest*, the weighing on those features is likely to reverse. However, when the words appear in the context of a sentence such as *The team lost the football in the forest*, the context might bring up more unusual features like 'Landmark', 'Fearful', and 'Surprise'. Thus, when words share features, those aspects of the word representation that are relevant to the context are strengthened (Hampton, 1996; Kiefer & Pulvermüller 2012; Medin & Shoben, 1988; Mitchell & Lapata, 2010; Murphy, 1990; Wisniewsky, 1998).

If this theory is correct, it should be possible to see such changes in the fMRI patterns of subjects that are reading words in different contexts. Such effect has indeed been demonstrated in earlier work (Aguirre-Celis & Miikkulainen, 2017, 2018, and 2019). Their model was able to identify the effect of similar context on different concepts (*boat crossed* vs. *car crossed*), as well as the effect of different contexts on the same concept (*bird flew* vs. *plane flew*). These effects were quantified across a large corpus of sentences, demonstrating that the meaning of the sentence context is transferred, to a degree, to each word in the sentence. The results were obtained by mapping sentence fMRI to the FGREP mechanism (Forming Global Representations with Extended BP, Miikkulainen & Dyer, 1991), to adjust those representations to take context into account.

What remains to be shown is that the changes in the word representation are actually meaningful to the subjects, i.e., that they are aware of them and agree on the predictions of the model. To that end, a human subject study is presented in this paper. Subjects were given words in context and asked to evaluate possible changes.

In the following sections, the modeling framework is first reviewed, including fMRI imaging data collection, the brain-based semantic model, and the neural network model that produces the predictions. The methods and results of the human subject study are then described, followed by the methods and results of the computational study. The methods and results of comparing the human judgements and the computational model predictions concludes the paper.

## Modeling Framework

The neural network model used in this study CEREBRA (Context-dependent mEaning REpresentation in the BRAin), was developed by Aguirre-Celis & Miikkulainen (2017, 2018, and 2019) to investigate how words change under the context of a sentence using imaging data. It is based on the CAR semantic feature model (Concept Attributes Representation, Binder, 2016) , and implemented using the FGREP neural network (Miikkulainen & Dyer, 1991). The model is trained to predict sentence fMRI, using CEREBRA to map CAR representations of words into fMRI data of subjects reading everyday sentences.

CARs (a.k.a. the experiential attribute representation model), represent the basic components of meaning defined in terms of known neural processes and brain systems. They are composed of a list of well-known modalities that correspond to specialized sensory, motor and affective brain processes, systems processing spatial, temporal, and casual information, and areas involved in social cognition. These aspects of mental experience model each word as a collection of a 66-dimensional feature vector that captures the strength of association between each neural attribute and the word meaning. For instance, Figure 1 shows the CAR for the concept *football*. For a more detailed account of the attribute selection and definition see Binder, et al. (2009, 2011, 2016a, and 2016b).

Aguirre-Celis & Miikkulainen's model was developed based on three sets of data: A sentence collection prepared by Glasgow et al. (2016), the semantic vectors (CAR ratings) for the words obtained via Mechanical Turk, and the fMRI images for the sentences, both collected by the Medical College of Wisconsin (Anderson, et al., 2016; Binder, et al., 2016). Additionally, fMRI representations for individual words (called SynthWord) were synthesized by averaging the sentence fMRI.

The CEREBRA model was trained to map the CAR representations of words in each sentence into the observed fMRI of the sentence (Figure 2). Gradient descent was then continued separately for each sentence, reducing the error by modifying only the CARs at the input of the network (i.e., using the FGREP method). As a result, the strengths of the attributes in the CARs changed according to how relevant each attribute is for that sentence context.

The CEREBRA model was trained 20 times for each of the eleven fMRI subjects with different random seeds. A total of 20 different sets of 786 context word representations (one word representation for each sentence where the word appears) were thus produced for each subject. Afterwards, the mean of the 20 representations was used as the final representation for each word. These context-based word vectors were then used as predictions for the human judgements obtained in the human subject study in this paper.

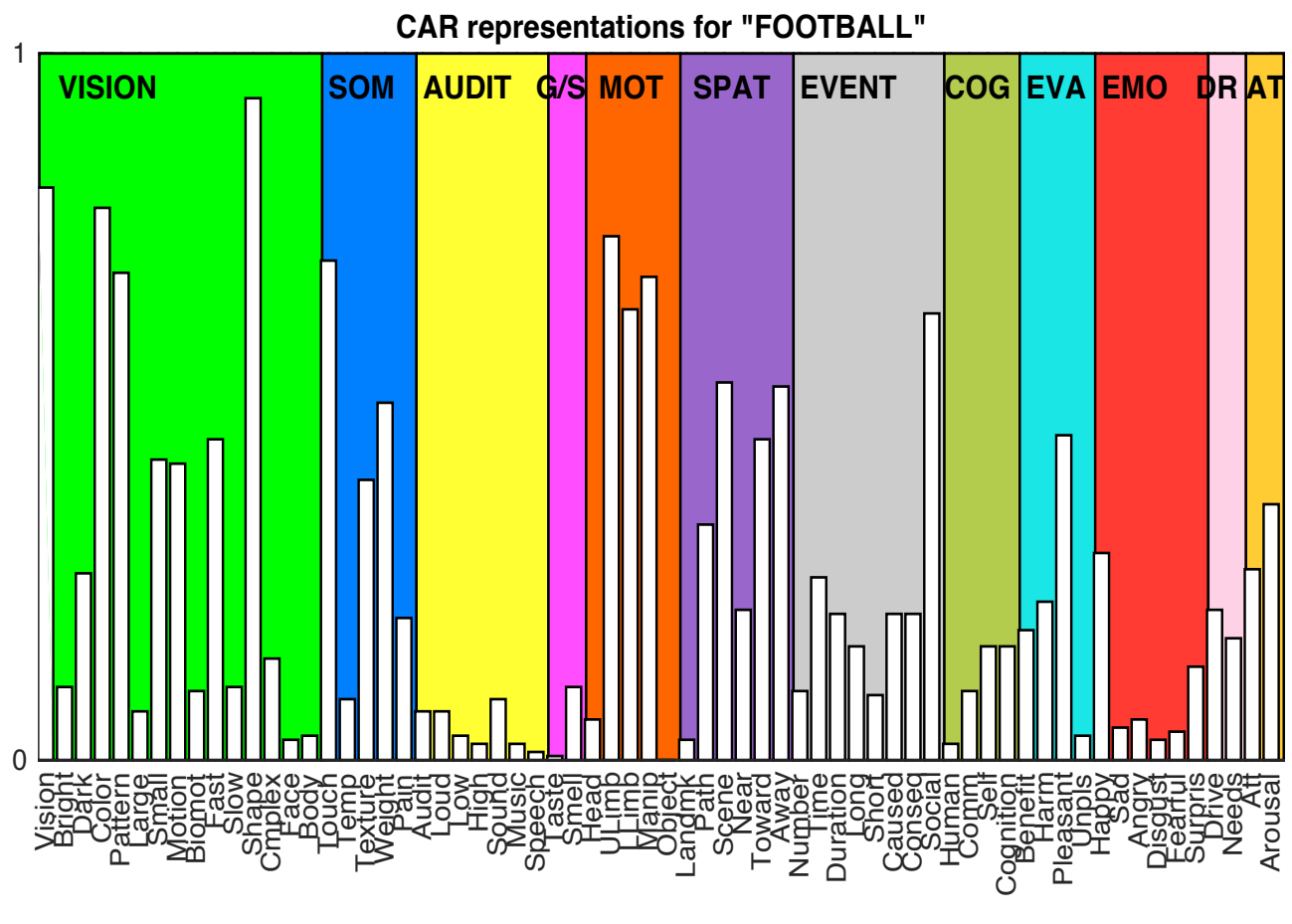

Figure 1: Bar plot of the 66 semantic features for the concept *football* (Binder, et al., 2016). The values represent average human ratings for each feature. Given that *football* is an object, it gets low weightings on human-related attributes such as Temperature, Speech, and Taste, and emotions including Sad and Angry, and high weightings on attributes like Shape, Touch, Lower-limb, and Manipulation.

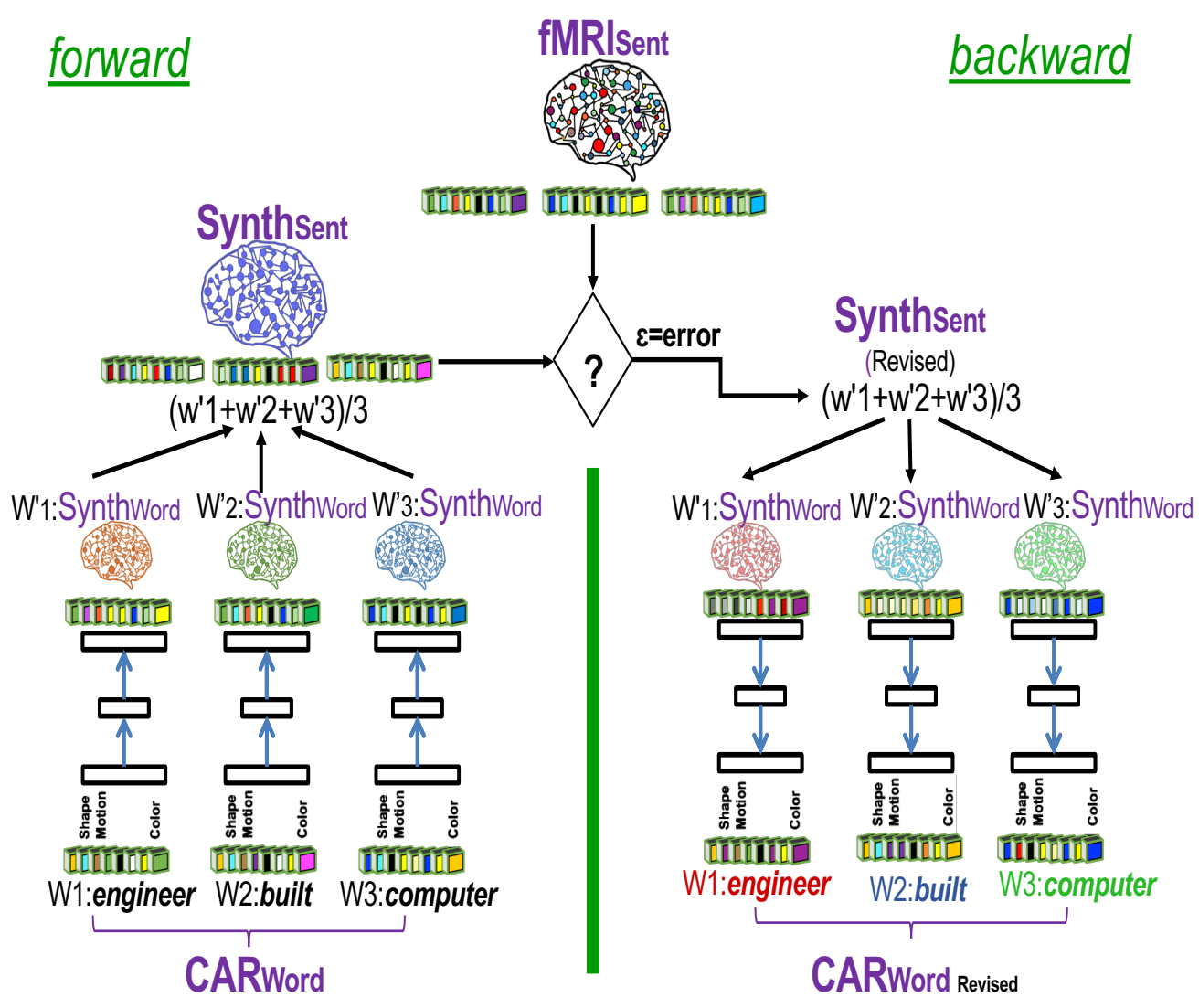

Figure 2: The CEREBRA model to account for context effects (Aguirre-Celis & Miikkulainen, 2017, 2018, and 2019). (1) Propagate CARWords to SynthWords. (2) Construct SynthSent by averaging the SynthWords into a prediction of the sentence. (3) Compare SynthSent against observed fMRI sentence. (4) Backpropagate the error with FGREP for each sentence, freezing network weights and changing only CARWords. (5) Repeat until error reaches zero or CAR components reach their upper or lower limits. Thus, the CEREBRA model captures context effects by mapping brain-based semantic representations to fMRI sentence images.

## Measuring Human Judgements

The purpose of the survey is to evaluate the computational model predictions addressing the central question: How does the meaning of a word change in different sentences? According to Aguirre-Celis & Miikkulainen (2017, 2018, and 2019), different attributes of the target word are weighted differently depending on context. Thus, the model is used to determine how the generic meaning of a word would have to change in order to account for the context. Specifically, the survey was designed to characterize these changes by asking the subject directly: In this context, how does this attribute change?

### Materials and Design

The survey design was based on the fMRI subject data and sentence collection, the CEREBRA predictions, and the CAR's literal descriptions. A script was implemented to select the most representative subjects, sentences, words, and word attributes. To make the questions more understandable for the participants, the original descriptions of the 66 attributes were rephrased to make the questionnaires easy to read and to respond to, while retaining the meaning of the original descriptions elaborated by Binder et al. (2016).

The data from the aggregation analysis prepared by Aguirre-Celis & Miikkulainen (2019), was used as the starting point, and filtered further to make it systematic and uniform. Only the centroid non-copula sentences, three word classes, and the top 10 statistically significant attribute changes for the target words (classes) were used. The final stimuli that met this criteria consisted of 64 different sentences from the Glasgow corpora containing the roles of Agent, Verb, and Patient/Object/Location/Event (POLE). Altogether contained 123 words: 38 Agents, 39 Verbs, and 46 POLE words. Table 1 shows a sample of 5 sentences and 15 words from this collection. The table on the top part, shows the original sentence number from the Glasgow collection, the sentence itself, and the number of times each particular sentence was selected by the script. The table on the bottom part, lists the three classes, the words in alphabetical order, and the word number from the original collection. Red indicates words used in two different roles in various sentences (e.g., *activist* as Agent or Patient).

Table 1: A sample of 5 sentences and a sample of 15 words used in the questionnaires are shown. The top part of the table shows the original sentence number, the sentence itself, and the number of times each sentence was included in the survey. The bottom part shows the words in alphabetical order divided into Agent, Verb, and any of Patient/Object/Location/Event (POLE). Words shown in red appeared in two different roles in separate sentences.

| Questionnaires Unique Sentences | | |
|---|---|---|
| No. | Sentence | Occ |
| 113 | *The author kicked the desk* | 4 |
| 116 | *The injured horse slept at night* | 1 |
| 149 | *The banker watched the peaceful protest* | 1 |
| 150 | *The mob approached the embassy* | 2 |
| 154 | *The politician celebrated at the hotel* | 2 |

| Questionnaires Unique Words | | | | | |
|---|---|---|---|---|---|
| No. | Agent | No. | Verb | No. | POLE |
| 2 | *activist* | 10 | *arrested* | 2 | *activist* |
| 13 | *author* | 12 | *ate* | 21 | *bird* |
| 15 | *banker* | 27 | *bought* | 25 | *boat* |
| 21 | *bird* | 31 | *broke* | 26 | *book* |
| 25 | *boat* | 38 | *celebrated* | 41 | *chicken* |

The complete survey is an array of 24 questionnaires that include 15 sentences each. For each sentence, the survey measures 10 attribute changes for each target word. Overall, each questionnaire thus contains 150 evaluations. For example, a questionnaire might measure changes on 10 specific attributes such as 'is visible', 'living thing that moves', 'is identified by sound', 'has a distinctive taste', for a specific word class as in *politician*, for 15 sentences such as *The politician celebrated at the hotel*. An example sentence questionnaire is shown in Figure 3.

To select which attributes to test the following process was used: (1) use the sentences with at least 10 statistically significant attribute changes (ssa), (2) from the 25 attributes with the largest change (or the number of ssa available), randomly select 10 within a sentence, and (3) organize the attribute collection for each question using Binder's (2016) original list arrangement.

The statistically significant attribute changes thus selected represent meaningful differences between the new and the original CAR representations. The point of the random selection within the top 25 was that: (1) there is a large number of potentially meaningful attributes, i.e. 25 at least; (2) for simplicity, the survey must not contain many questions; (3) the differences among the top 25 are not very large; and (4) it is necessary to get a varied selection of attributes. Choosing the top 10 instead would have resulted in too many visual features for most sentences, either because they frequently changed more, or because visual attributes are more numerous (i.e., 15 out of the 66).

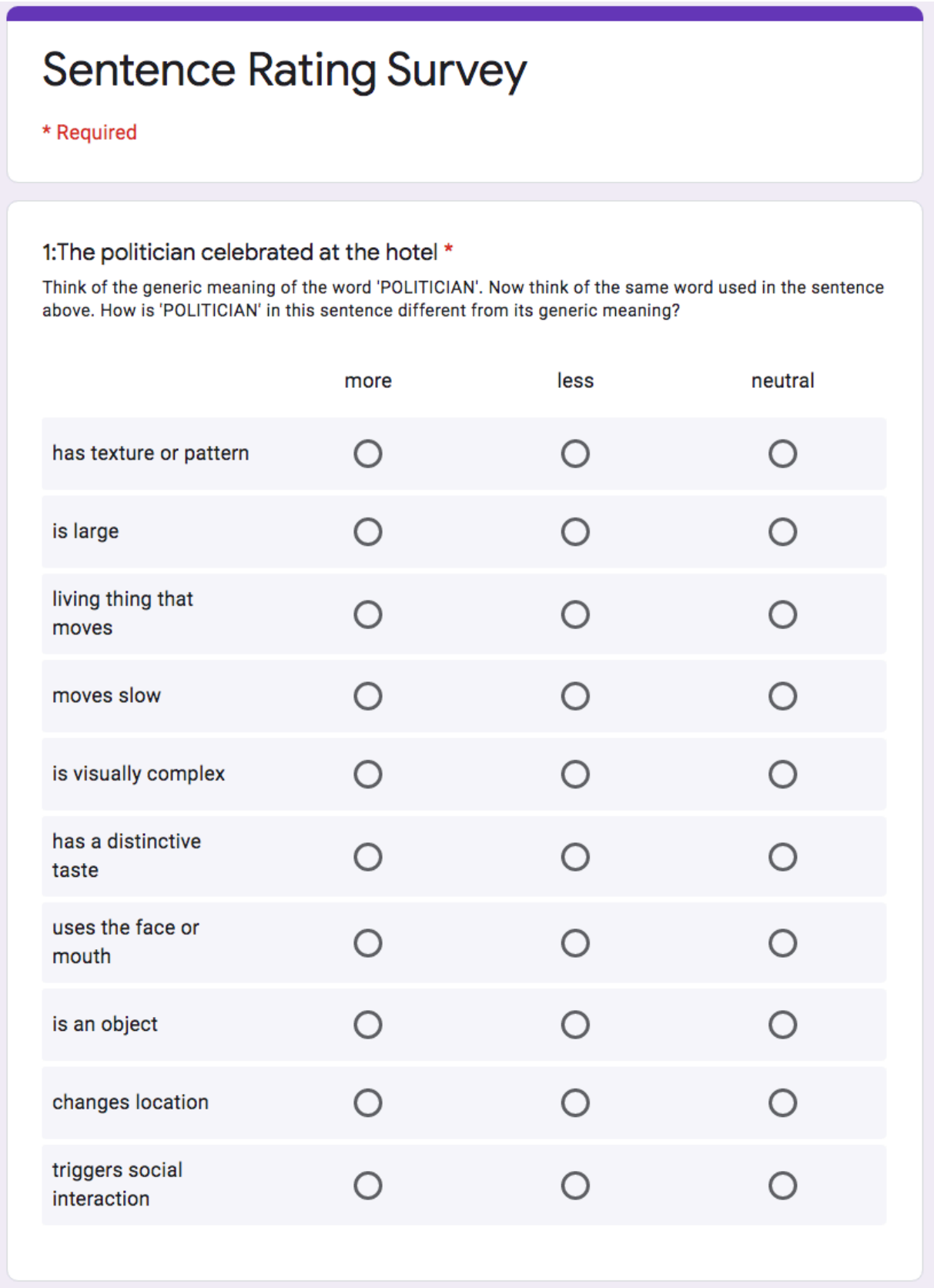

Figure 3: Example sentence in a questionnaire prepared to evaluate the computational model results. The sentence is *The politician celebrated at the hotel*, the target word is *politician* in the role of Agent. Ten different attribute changes are measured by selecting whether the attribute increased ("more"), decreased ("less") or remained "neutral". The human judgements were thus matched with those predicted by the CEREBRA model trained with the fMRI data.

## Participants

Human judgements were crowdsourced using Google Forms in accordance with the University of Texas Institutional Review Board (2018-08-0114). The experiments were completed by 27 unpaid volunteers (nine females). The participants' ages ranged from 18 to 64 years, with the mean of 33. Nineteen of them were self-reported bilinguals (English as a second language) and eight English native speakers. Four subjects were affiliated with The University; the rest of the population consisted of working people residing in different parts of north and central America (Texas, Seattle, California, Costa Rica, and Mexico). The subjects had no background in linguistics, psychology or neurosciences.

## Procedure

The 24 questionnaires were designed using Google Forms. The respondents were asked to think how the meaning of a specific word changes within the context of a sentence compared to its generic meaning, by evaluating which word attributes change "more", "less", or stay the same.

Subjects were recruited by sending emails or text messages directly along with the survey link to access their assigned questionnaire. The data collection was done online and the participants responded using their cell phone or personal computer. Each questionnaire consisted of an Introduction, Description of the Experiment, Example, and the Survey. Each questionnaire takes about 15 minutes to complete.

Three of the participants responded to all of the 24 questionnaires. The entire survey consisted of a total of 3600 questions, so it took them four to seven days to complete this task at a pace of approximately four questionnaires (i.e., an hour per day). This task was a lot of work, the fourth set of responses was obtained by distributing it among multiple raters: twenty-four additional participants were recruited to each respond to one of the 24 questionnaires.

Table 2: Distribution analysis and inter-rater agreement. The top part shows human judgement distribution for the three possible questionnaire responses "less" (-1), "neutral" (0), and "more" (1). The bottom part shows percent agreement for the four raters. The task was difficult and the agreement low. Only those questions where three out of four participants agreed were considered reliable and compared to the CEREBRA model.

| **HUMAN RESPONSES DISTRIBUTION** | | | | | | |
|---|---|---|---|---|---|---|
| **Resp/Part** | **P1** | **P2** | **P3** | **P4** | **AVG** | **%** |
| -1 | 2065 | 995 | 645 | 1185 | 1223 | 34.0% |
| 0 | 149 | 1120 | 1895 | 1270 | 1109 | 30.8% |
| 1 | 1386 | 1485 | 1060 | 1145 | 1269 | 35.3% |
| TOT | 3600 | 3600 | 3600 | 3600 | 3600 | 100% |
| **PARTICIPANT AGREEMENT ANALYSIS** | | | | | | |
| | **P1** | **P2** | **P3** | **P4** | **AVERAGE** | **%** |
| **P1** | 0 | 1726 | 1308 | 1650 | 1561 | 43% |
| **P2** | 1726 | 0 | 1944 | 1758 | 1809 | 50% |
| **P3** | 1308 | 1944 | 0 | 1741 | 1664 | 46% |
| **P4** | 1650 | 1758 | 1741 | 0 | 1716 | 48% |
| | | | | | | |
| | | | | **TOTAL** | **6751** | |
| | | | | **AVG xPAR** | **1688** | |
| | | | | | | |
| | | **AVERAGE** | Particip match each other | | | 47% |

## Results

Human responses were first characterized through data distribution analysis. Table 2 shows the number of answers "less" (-1), "neutral" (0), and "more" (1) for each respondent. Columns labeled P1, P2, and P3, show the responses of the three participants that were assigned the entire survey (24 questionnaires, 3600 answers). Column labeled P4 shows the combined answers of the 24 different participants responding to one questionnaire each. The top part of the table shows the distribution of the rater's responses and the bottom part shows the level of agreement among them. As can be seen, participants agreed only 47% of the time.

According to Grand, et. al (2018) it is not worth comparing system predictions vs. human judgements if inter-subject reliability is too low. However, since there were a lot of questions, it was possible to include only questions that were the most reliable, i.e., where three out of four participants agreed. There were 1966 such questions or 55% of the total set of questions.

# Measuring Model Predictions

Three different approaches were designed to quantify the predictions of the FGREP model. In order to measure the level of agreement between humans and FGREP a model fitting procedure was implemented.

## Quantifying the FGREP Predictions

The survey directly asks for the direction of change of a specific word attribute in a particular sentence, compared to a generic meaning. Since the changes in the CEREBRA model range within (-1,1), in principle that is exactly what the model produces. However, Aguirre-Celis & Miikkulainen (2017, 2018, and 2019) found that some word attributes always increase, and do so more in some contexts than others. This effect is related to conceptual combination (Hampton, 1996; Wisniewsky, 1998), contextual modulation (Barclay, 1974), or attribute centrality (Medin & Shoben, 1988): the same property is true for two different concepts but more central to one than to the other (e.g., it is more important for boomerangs to be curved than for bananas).

The direction of change is therefore not a good predictor of human responses; instead these changes need to be measured relative to changes in other words. Thus such approaches were evaluated:

1. What is the effect of the rest of the sentence in the target word? This effect was measured by computing the average of the CEREBRA changes (i.e., new-original) of the other words in the sentence, and subtracting that average change from the change of the target word:

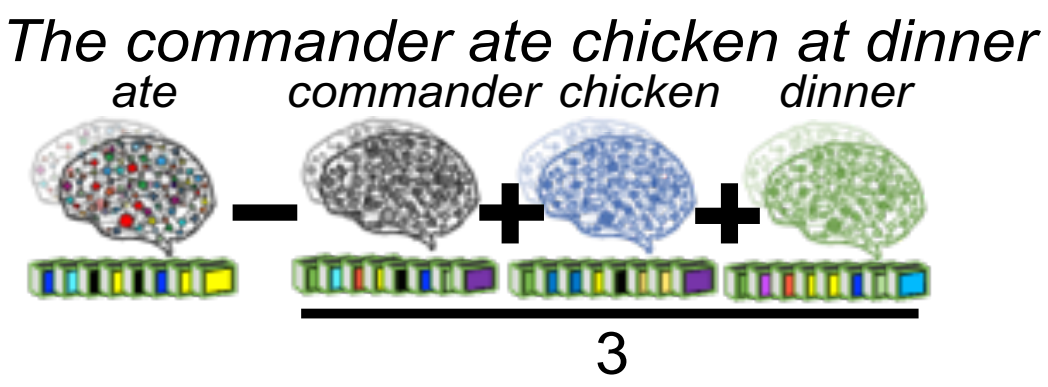

2. What is the effect of the entire sentence in the target word? This effect was measured by computing the average of the CEREBRA changes (i.e., new-original) of all the words in the sentence including the target word, and subtracting that average change from the change of the target word:

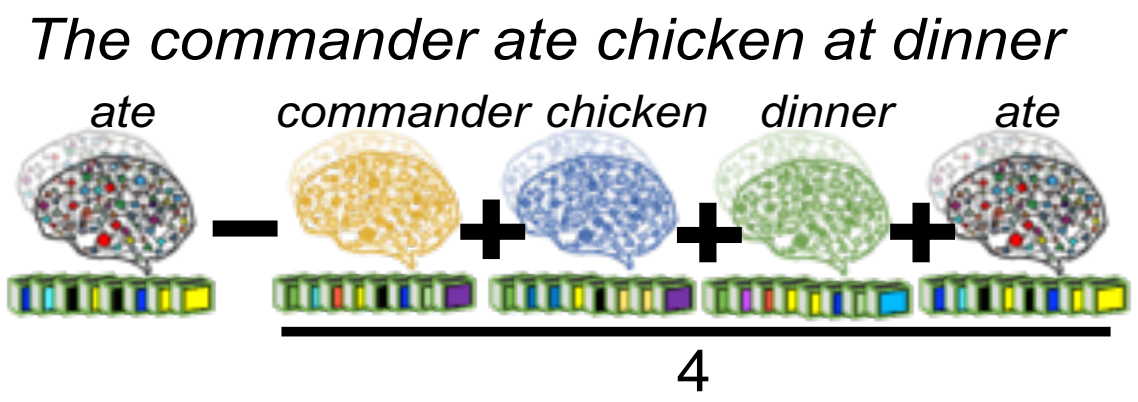

3. What is the effect of CARs used in context as opposed to CARs used in isolation? This effect was measured by computing the average of the CEREBRA changes (i.e., new-original) of the different representations of the same word in several contexts, and subtracting that average change from the change of the target word:

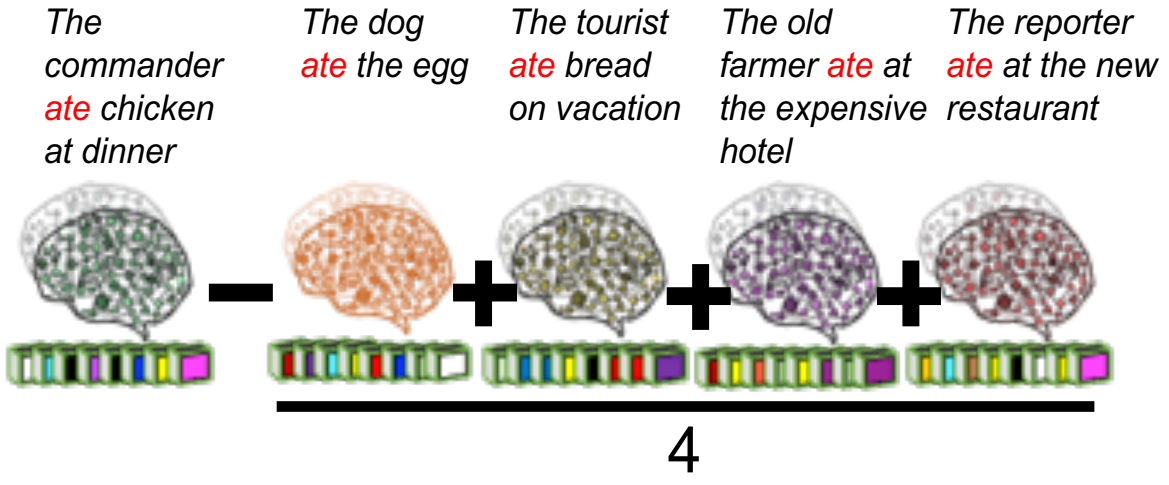

The first two approaches have the advantage of being simple. However the third approach is motivated by neurological evidence suggesting that sentence comprehension involves a common core representation of multiple word meanings combined into a network of regions distributed across the brain (Anderson, et al., 2016; Gennari, et. al., 2007). In line with this view, a generic (or isolated) word representation can be formed by averaging the activity in multiple sentence contexts.

In each of these cases, the resulting vectors are expected to accurately represent the direction of change asked in the questionnaires. They are the ratings used in the evaluation procedure described in the following section.

## Procedure

Starting from a different random seed, the CEREBRA model was trained 20 times for each of the eight best fMRI subjects (i.e., where the fMRI data in general was most consistent). Responses for each model where thus obtained for the 1966 questions where three out of four participants agreed. In order to demonstrate that the CEREBRA model has captured human performance, the agreements of the CEREBRA changes and human surveys need to be at least above chance. Therefore a baseline model that generated random changes in the same range as the CEREBRA model was created. The chance model was queried 20 times for each of the 1966 questions, for each of the eight subjects. In this manner, 20 means and variances for each of the eight subjects for both CEREBRA and chance were created.

To estimate the level of agreement of CEREBRA and chance models with humans, a single parameter in each model was fit to human data: the boundary value above which the change was taken to be an increase (i.e., "more") or decrease/no change (i.e., "less"/"neutral"). The "less" and "neutral" categories were combined because they were much smaller than the "more" category in human data. The optimal value for this parameter was found by simply sweeping through the range (-1..1) and finding the value that measured on the highest number of matching responses where the 1966 questions are.

Table 3: Matching CEREBRA predictions with human data, compared to chance. The table shows the average agreement of the 20 repetitions across all subjects. CEREBRA agrees with human responses 54% when the chance level is 45%.

| PARTICIPANTS AVERAGE AGREEMENT | | | |
|---|---|---|---|
| RATINGS | HUMAN | CEREBRA | CHANCE |
| -1/0 | 1074 | 466 | 8 |
| 1 | 892 | 587 | 886 |
| TOTAL | 1966 | 1052 | 894 |
| | AVERAGE | 54% | 45% |

## Matching Predictions with Human Judgements

The three approaches to measuring the predictions of the CEREBRA model, i.e., the context effect of the rest of the sentence, the context effect of the entire sentence, and the context effect of the word in different contexts, were implemented and fit to human data using single-boundary model fitting. The three sets of data produced very similar results, therefore only those of the third approach, are reported in this paper. In fact, the other two approaches achieved slightly better results than this one (by 1%).

The match results are presented in Table 3 and the statistical significance in Table 4. The CEREBRA model matches human responses in 54% of the questions when the chance level is 45% - which is indistinguishable from always guessing "more", i.e., the largest category of human responses. The differences shown in Table 4 are statistically strongly significant for all of the eight subjects. These results show that the changes in word meanings due to sentence context that are observed in the fMRI and interpreted to semantic feature representations are real and meaningful to the subjects.

Table 4: Statistical analysis for CEREBRA and chance. The table shows the means and variances of CEREBRA and chance models for each subject and the *p*-values of the *t*-test, showing that the differences are highly significant.

| SUBJECTS | CEREBRA | | CHANCE | | p-value |
|---|---|---|---|---|---|
| | MEAN | VAR | MEAN | VAR | |
| S1 | 1033 | 707.25 | 894 | 6.01 | 3.92E-24 |
| S2 | 1035 | 233.91 | 894 | 7.21 | 6.10E-33 |
| S3 | 1063 | 224.41 | 894 | 11.52 | 5.22E-36 |
| S4 | 1077 | 94.79 | 894 | 7.21 | 3.89E-44 |
| S5 | 1048 | 252.79 | 895 | 12.03 | 1.83E-33 |
| S6 | 1048 | 205.82 | 894 | 4.62 | 1.73E-35 |
| S7 | 1075 | 216.77 | 895 | 7.21 | 1.65E-37 |
| S8 | 1039 | 366.06 | 894 | 2.52 | 6.10E-30 |

## Discussion and Future Work

The study provides a missing piece on the theory of semantic feature representations: The context-dependent changes in them are actionable and can be used to predict human judgements. Given how noisy human responses data is, the 9% difference between CEREBRA and chance is a strong result.

An interesting direction for future work would be to replicate the study on a more extensive data set with a fully balanced stimuli and with fMRI images of individual words. The differences should be even stronger and should be possible to uncover even more refined effects. Such data should also improve the survey, since it would be possible to identify questions where the effects can be expected to be more reliable. Inter-raters reliability could also be improved by training the raters better so that they are comfortable with the concept of generic meaning and the concept of variable meanings. It may also be possible to design the questions such that they allow comparing alternatives which may be easier for the participants.

Compared to other approaches, such as distributional semantic models (DSMs), CAR theory enables direct correspondence between conceptual content and neural representations that engage when instances of the concept are experienced. In contrast, DSMs are not grounded on perception and motor mechanisms, instead their representations reflect the semantic knowledge acquired through a lifetime of linguistic experience. The richness and complexity of the representations in the CAR theory offers a powerful method to further explore the semantic space. To this end, the CEREBRA model (Aguirre-Celis & Miikkulainen, 2017, 2018, 2019) have demonstrated that CARs can capture fine distinctions in meaning, creating many possibilities of improvements of the theory itself, and supporting the explainable feature representations of CARs by discerning how concepts are dynamically represented in the brain. DSM representations inherently cannot be interpreted.

## Conclusion

This paper provides experimental and computational support on these main ideas: (1) context-dependent meaning representations are embedded in the fMRI sentences, and (2) they can be characterized. Using brain-based semantic feature representations (CARs) together with the CEREBRA change model, (3) such changes are real and meaningful to the subjects. It therefore takes a step towards understanding how the brain constructs sentence-level meanings from word-level attributes.

## Acknowledgments

We would like to thank Jeffery Binder (Medical College of Wisconsin), Rajeev Raizada and Andrew Anderson (University of Rochester), Mario Aguilar and Patrick Connolly (Teledyne Scientific Company) for their work and valuable help regarding this research. This work was supported in part by IARPA-FA8650-14-C-7357 and by NIH 1U01DC014922 grants.

## References

Aguirre-Celis, N., Miikkulainen R. (2017). From Words to Sentences & Back: Characterizing Context-dependent Meaning Rep in the Brain. *Proceedings of the 39th Annual Conference of the Cognitive Science Society*, London, UK. 1513-1518.

Aguirre-Celis N., Miikkulainen R. (2018) Combining fMRI Data and Neural Networks to Quantify Contextual Effects in the Brain. In: Wang S. et al. (Eds.). *Brain Informatics*. BI 2018. Lecture Notes in Computer Science. 11309, 129-140. Springer, Cham.

Aguirre-Celis, N., Miikkulainen R. (2019). Quantifying the Conceptual Combination Effect on Words Meanings. *Proceedings of the 41th Annual Conference of the Cognitive Science Society*, Montreal, CA. 1324-1331.

Anderson, A. J., Binder, J. R., Fernandino, L., Humpries C. J., Conant L. L., Aguilar M., Wang X., Doko, S., Raizada, R. D. (2016). Predicting Neural activity patterns associated with sentences using neurobiologically motivated model of semantic representation. *Cerebral Cortex*. 1-17.

Barclay, J.R., Bransford, J.D., Franks, J.J., McCarrell, N.S., & Nitsch, K. (1974). Comprehension and semantic flexibility. *Journal of Verbal Learning and Verbal Behavior, 13*, 471–481.

Binder, J. R., Desai, R. H., Graves, W. W., Conant L. L. (2009). Where is the semantic system? A critical review and meta-analysis of 120 functional neuroimaging studies. *Cerebral Cortex*, 19:2767-2769.

Binder, J. R., Desai, R. H. (2011). The neurobiology of semantic memory. *Trends in Cognitive Science*, 15(11), 527-536.

Binder, J. R., Conant L. L., Humpries C. J., Fernandino L., Simons S., Aguilar M., Desai R. (2016a). Toward a brain-based componential semantic representation. *Cognitive Neuropsychology,* 33(3-4), 130-174.

Binder, J. R. (2016b). In defense of abstract conceptual representations. *Psychonomic Bulletin & Review*, 23.

Gennari,S., MacDonald, M., Postle, B., Seidenberg, S. (2007). Context-dependent interpretation of words: Evidence for interactive neural processes. *NeuroImage*, 35(3), 1278-1286.

Glasgow, K., Roos, M., Haufler, A. J., Chevillet, M., A., Wolmetz, M. (2016). Evaluating semantic models with word-sentence relatedness. arXiv:1603.07253.

Grand, G., Blank, I., Pereira, F., Fedorenko, E. (2018). Semantic Projection: Recovering Human Knowledge of Multiple, Distinct Object Features from Word Embeddings. arXiv:1802.01241v2

Hampton, J. (1997). Conceptual combination. In K. Lamberts & D. R. Shanks (Eds.), *Studies in cognition. Knowledge, concepts and categories*. MIT Press 133–159.

Kiefer M, Pulvermüller F. (2012). Conceptual representations in mind and brain: theoretical developments, current evidence and future directions. *Cortex*, 48, 805–825.

Medin, D. L., & Shoben, E. J. (1988). Context and structure in conceptual combination. *Cognitive Psychology*, 20, 158-190.

Mitchell J, Lapata M. (2010). Composition in distributional models of semantics. *Cognitive Science* 38(8), 1388–1439.

Miikkulainen, R., Dyer, M., G. (1991). Natural Language Processing with Modular PDP Networks and Distributed Lexicon. *Cognitive Science*, 15, 343-399.
Murphy, G. (1990). Noun Phrase Interpretation and Conceptual Combination. *Journal of Memory and Language*, 29, 259-288.
Wisniewski, E., Property Instantiation in Conceptual Combination. (1998). *Memory & Cognition*, 26, 1330-1347.